\documentclass[preprint,12pt]{elsarticle}

\usepackage{amssymb}
\usepackage{amsmath,amsfonts}
\usepackage{algorithmic}
\usepackage{algorithm}
\usepackage{array}
\usepackage[caption=false,font=normalsize,labelfont=sf,textfont=sf]{subfig}
\usepackage{textcomp}
\usepackage{stfloats}
\usepackage{url}
\usepackage{verbatim}
\usepackage{graphicx}
\usepackage{multirow}
\usepackage{bm}
\usepackage{color}

\usepackage{booktabs}
\usepackage{tabularx}
\usepackage{threeparttable}

\journal{Applied Soft Computing}

\begin{document}

\begin{frontmatter}



\title{PeriodNet: Boosting the Potential of Attention Mechanism for Time Series Forecasting} 


\author[swjtu,ti]{Bowen Zhao}
\author[swjtu,ti]{Huanlai Xing\corref{cor1}} \ead{hxx@home.swjtu.edu.cn}
\author[swjtu,ti]{Zhiwen Xiao}
\author[swjtu,ti]{Jincheng Peng}
\author[swjtu,ti]{Li Feng}
\author[swjtu,ti]{Xinhan Wang}
\cortext[cor1]{Corresponding author}
\affiliation[swjtu]{%
	organization={School of Computing and Artificial Intelligence, Southwest Jiaotong University},
	city={Chengdu},
	country={China},
}
\affiliation[ti]{organization={Tangshan Institute of Southwest Jiaotong University},
	city={Tangshan},
	country={China}}

\author[uon]{Rong Qu}
\affiliation[uon]{%
	organization={School of Computer Science, University of Nottingham},
	city={Nottingham},
	country={UK},
}

\author[xju]{Hui Li}
\affiliation[xju]{%
	organization={School of Mathematics and Statistics, Xi’an Jiaotong University},
	city={Xi’an},
	country={China},
}

\begin{abstract}
	The attention mechanism has demonstrated remarkable potential in sequence modeling, exemplified by its successful application in natural language processing with models such as Bidirectional Encoder Representations from Transformers (BERT) and Generative Pre-trained Transformer (GPT). Despite these advancements, its utilization in time series forecasting (TSF) has yet to meet expectations. Exploring a better network structure for attention in TSF holds immense significance across various domains. In this paper, we present PeriodNet with a brand new structure to forecast univariate and multivariate time series. PeriodNet incorporates period attention and sparse period attention mechanism for analyzing adjacent periods. It enhances the mining of local characteristics, periodic patterns, and global dependencies. For efficient cross-variable modeling, we introduce an iterative grouping mechanism which can directly reduce the cross-variable redundancy. To fully leverage the extracted features on the encoder side, we redesign the entire architecture of the vanilla Transformer and propose a period diffuser for precise multi-period prediction.
	Through comprehensive experiments conducted on eight datasets, we demonstrate that PeriodNet outperforms six state-of-the-art models in both univariate and multivariate TSF scenarios in terms of mean square error and mean absolute error. In particular, PeriodNet achieves a relative improvement of 22\% when forecasting time series with a length of 720, in comparison to other models based on the conventional encoder-decoder Transformer architecture.
\end{abstract}

\begin{highlights}
	\item The proposed PeriodNet hybridizes a period attention mechanism, an iterative grouping mechanism, and a period diffuser architecture to achieve accurate multivariate time series forecasting.
	\item The period attention mechanism captures temporal similarities among adjacent periods to improve time series modeling.
	\item The iterative grouping mechanism jointly analyzes time series with similar characteristics, enabling cross-variate modeling with low redundancy.
	\item The period diffuser architecture leverages multi-scale period features extracted by the encoder to enhance the accuracy and efficiency of time series forecasting.
\end{highlights}

\begin{keyword}
	Attention Mechanism\sep Data Mining\sep Deep Learning\sep Time Series Forecasting
	
	
\end{keyword}

\end{frontmatter}



\section{Introduction}
Among various data modalities, time series data have gained immense significance across a growing number of domains, such as industrial applications , financial sectors \citep{LI2025113241}, and medical fields \citep{ASHOFTEH2022109422}. Time series forecasting (TSF), a valuable data mining task, involves analyzing temporal patterns within historical data to make predictions about future processes \citep{WANG2025113779}.

In the pursuit of accurate forecasting, locality, periodicity, and temporal redundancy are critical characteristics that have to be considered. Locality \citep{liEnhancingLocalityBreaking2020,nieTimeSeriesWorth2023,zhangCROSSFORMERTRANSFORMERUTILIZING2023} is a widely used concept indicating that patterns in time series are influenced more by neighboring information than by distant one. These patterns tend to exhibit certain dependencies or relationships within specific temporal ranges or intervals. Disregarding the characteristics of locality may result in severe performance degradation. 
Periodicity \citep{oreshkinNBEATSNeuralBasis2020} refers to the recurring pattern or behavior that repeats at regular intervals over time. It enables the identification and utilization of recurring patterns, contributing to the development of more accurate predictive models.	
As for the temporal redundancy \citep{liMTSMixersMultivariateTime2023}, it is relevant to the sampling rate when collecting time series and refers to the presence of unnecessary or repetitive information that may not contribute substantially to the underlying patterns. The existence of redundancy in a time series can affect TSF by potentially introducing noise interference.
The remarkable progress in deep learning (DL) enables the effective analysis of these features and has significantly influenced the performance of TSF in recent years.

As one of the most widely embraced DL models, Transformer \citep{vaswaniAttentionAllYou2017} exhibits remarkable performance across a variety of domains. Its variants, such as Vision Transformer \citep{dosovitskiyImageWorth16x162021}, Swin Transformer \citep{liuSwinTransformerHierarchical2021}, and DAT \citep{xiaVisionTransformerDeformable2022}, have been proposed to achieve state-of-the-art results in computer vision (CV) tasks. Furthermore, Transformer has proven its effectiveness as a foundational technique in models like BERT \citep{devlinBERTPretrainingDeep2019}, Poolingformer \citep{zhangPoolingformerLongDocument2021}, and GPT-4 \citep{openaiGPT4TechnicalReport2023} in natural language processing (NLP). In the realm of TSF, there have been numerous attempts to leverage Transformer-based algorithms. LogSparse Transformer and Informer have made notable progress benefiting from global awareness. However, the deficiency of local information makes them insensitive in capturing short-period patterns. Autoformer \citep{wuAutoformerDecompositionTransformers2021} and FEDformer \citep{zhouFEDformerFrequencyEnhanced2022}, inheriting the Encoder-Decoder structure of Transformer, mitigate noise interference by employing deep decomposition architectures and analyzing features in the frequency domain, resulting in lower prediction errors. Nonetheless, recently proposed algorithms, such as DLinear \citep{zengAreTransformersEffective2022}, MTS-Mixers \citep{liMTSMixersMultivariateTime2023}, and SparseTSF \citep{lin2024sparsetsf}, based on linear prediction, exhibit surprisingly superior performance with much lower complexity compared to the aforementioned Transformer-based models. Therefore, typical Transformers in TSF are not that effective and boosting the potential of attention offers a means for extracting sufficient and multi-scale temporal features.

In multivariate TSF, effective cross-variable modeling poses a significant challenge that hampers prediction accuracy. Many existing attention-based algorithms address this by combining the multivariate time series (MTS) in the embedding stage and treating them similarly to univariate data. This approach offers computational and memory savings compared to methods like the channel independence assumption in DLinear \citep{zengAreTransformersEffective2022} and explicit cross-variable modeling in Crossformer \citep{zhangCROSSFORMERTRANSFORMERUTILIZING2023} and CA-SFCN \citep{haoNewAttentionMechanism2020}. However, the approach above is not capable of effectively differentiating between distinct features originating from multiple variables, resulting in non-trivial errors in multivariate TSF.

In the vanilla Transformer, decoders rely on the final output of the encoder for cross attention. Recent advancements incorporate linear prediction directly after the corresponding encoders. Nevertheless, both structures can only make predictions based on global features from the last layer, disregarding the valuable details accumulated in lower layers. Moreover, the prediction errors are significantly influenced by the input length of the traditional decoder, an additional hyperparameter, making models harder to train. That is why we doubt the suitability of the vanilla Transformer architecture for TSF.

In order to address the aforementioned challenges, we present a novel TSF-specific Transformer called PeriodNet, which specifically targets the locality, periodicity, and redundancy presented in time series data. PeriodNet introduces two enhanced temporal token mixers to effectively capture cross-time dependencies. It incorporates an iterative grouping mechanism to facilitate efficient modeling of MTS with reduced complexity. Moreover, PeriodNet optimizes the vanilla Transformer architecture and proposes a period diffuser to boost the potential of Transformer in TSF.
Our contributions are summarized as follows:
\begin{itemize}
	\item We introduce two novel mechanisms in our approach: period attention and sparse period attention. The period attention mechanism analyzes time series data from the period perspective and queries the similarity among adjacent periods to capture relevant temporal information. Based on the period attention, the sparse period attention mechanism concentrates on the same position within each adjacent period, better reducing the impact of temporal redundancy.
	\item We introduce an innovative iterative grouping mechanism to facilitate MTS modeling, where MTS data are grouped by learnable parameters based on the similarities between them, like frequency and trend. This allows for efficient and low-complexity modeling, enabling cross-variable period-wise temporal analysis within each group.
	\item In order to boost the performance of the attention mechanism in TSF, we for the first time introduce a novel TSF-specific architecture, called period diffuser. This architecture enables joint analysis of MTS data at the encoder side. Besides, it realizes separate predictions for each time series starting from longer periods and gradually moving towards shorter periods at the diffuser side.
	\item Simulation results demonstrate that the proposed PeriodNet outperforms six state-of-the-art (SOTA) algorithms on eight well-known multivariate TSF datasets and achieves excellent performance on four commonly used univariate TSF datasets in terms of the mean square error (MSE) and the mean absolute error (MAE).
\end{itemize}

\section{Related Work}
\subsection{Deep learning for time series forecasting} 
The mainstream TSF algorithms can roughly be classified into two categories: Transformer-based and traditional DL algorithms. LogSparse Transformer \citep{liEnhancingLocalityBreaking2020} is an early attempt for time series forecasting. It identifies redundant query vectors in the dot-product process and employs LogSparse self-attention to query only a subset of tokens for low complexity. LogSparse Transformer primarily focuses on low-rank global attention, resulting in lower complexity and prediction errors compared to traditional TSF models such as ARIMA \citep{ariyoStockPricePrediction2014} and DeepAR \citep{salinasDeepARProbabilisticForecasting2020}.
Another mainstream TSF method is based on the deep decomposition architecture, widely utilized in models like Autoformer \citep{wuAutoformerDecompositionTransformers2021} and FEDformer \citep{zhouFEDformerFrequencyEnhanced2022}. These models process seasonal information and trend-cyclical information individually. The importance of locality in TSF has gained considerable attention \citep{nieTimeSeriesWorth2023}, and only relying on global attention may limit the performance of Transformers. Inspired by the Vision Transformer \citep{dosovitskiyImageWorth16x162021}, PatchTST \citep{nieTimeSeriesWorth2023} incorporates patch-wise dot-product attention to capture local semantic information. However, it overlooks the fine-grained details within each patch, and the optimal number of patches varies depending on the length of the time series. Triformer presents a variable-specific patch attention to aggregate local features with linear complexity. It stacks attention layers in a triangular structure to gradually analyze global dependency \citep{cirsteaTriformerTriangularVariableSpecific2022}.
Pyraformer \citep{liuPYRAFORMERLOWCOMPLEXITYPYRAMIDAL2022a} proposes a pyramidal attention module based on a tree structure to flexibly extract temporal dependencies. Meanwhile, it utilizes coarse-scale construction module to facilitate information exchange among nodes in the multi-resolution tree.

There have been some effective non-Transformer DL approaches for TSF. Convolutional neural networks (CNNs), long short-term memory (LSTM) networks and fully connected networks (FCNs) are widely-used building blocks. 
LSTNet \citep{laiModelingLongShortTerm2018} extracts local dependencies among multiple variables through CNNs and mines long-term patterns based on LSTM networks. TPA-LSTM \citep{shihTemporalPatternAttention2019} uses CNNs to analyze time-invariant temporal patterns according to the hidden states from LSTM networks and uses attention mechanism to select TSF-related features. N-BEATS \citep{oreshkinNBEATSNeuralBasis2020} employs FCNs with residual connections to model the seasonality and trend in TSF. MTGNN \citep{wuConnectingDotsMultivariate2020} models relations in MTS from a graph-based perspective and applies graph neural networks (GNNs) to mine spatial and temporal relations for TSF. TimesNet \citep{wuTIMESNETTEMPORAL2DVARIATION2023} is a temporal variation model that incorporates Fast Fourier Transformation (FFT) with Inception blocks. MICN \citep{wangMICNMULTISCALELOCAL2023} introduces an isometric convolution block for global-local feature extraction.

Most of the existing studies overlook the interplay between adjacent periods and the interdependencies that exist across different periods. Moreover, it is difficult for these studies to effectively consolidate global and local semantic information. In light of these limitations, we propose the use of temporal token mixers integrated with a cross-variable mechanism to address these challenges above.

\subsection{Predictors in time series forecasting}
We refer to the structure that generates the target sequences in TSF as the predictor. In the vanilla Transformer and LogSparse Transformer, the predictors are decoders that generate sequences in an autoregressive manner. This approach involves using the current output as the input for the next prediction, resulting in inference speed being proportional to the length of the prediction.
To mitigate the error accumulation problem in autoregressive predictors, Adversarial Sparse Transformer \citep{wuAdversarialSparseTransformer} employs generative adversarial networks to improve prediction accuracy at the sequence level.
Informer \citep{zhouInformerEfficientTransformer2021a} introduces the ProbSparse attention mechanism to analyze the top-u queries. Based on this, it surpasses the limitations by proposing a generative-style decoder that can generate the entire target time series in a single forward process. Subsequent studies such as Non-stationary Transformer \citep{liuNonstationaryTransformersExploring} and Preformer \citep{duPreformerPredictiveTransformer} adopt similar predictors with different encoders. In addition, fully connected predictors have become popular.
DLinear \citep{zengAreTransformersEffective2022} is a simple FCN-based predictor integrated into a decomposition structure, showcasing the effectiveness and versatility of FCN-based predictors. This achievement motivates researchers to question whether Transformers are as powerful in TSF as they are in CV and NLP. Crossformer \citep{zhangCROSSFORMERTRANSFORMERUTILIZING2023} considers multi-scale features by generating series from small to large patches in encoder blocks and aggregating series from different patches for the final prediction. However, this approach may not well capture cross-patch relationships and is restricted by fixed-size patches. In TSF, accurately predicting both global-local dependencies and local details is crucial.

In summary, most existing predictors only utilize the final output of their encoders, significantly limiting the accuracy of TSF. Unlike data such as sentences and images, time series commonly exhibit strong locality and periodicity. Therefore, we redesign the structures of existing predictors and propose a multi-scale predictor called the period diffuser, which fully leverages the multi-period representations mined by the encoder to improve the TSF accuracy.

\section{Methodology}
As aforementioned, both local and global dependencies play a crucial role in TSF. To this end, we design a period attention mechanism to extract local features and a period router to adaptively capture global relations in a periodic perspective. Additionally, we incorporate a sparse period attention mechanism to eliminate temporal redundancy. To analyze cross-variable relationships, we develop an iterative grouping mechanism to collectively model MTS. In pursuit of precise forecasting, we introduce a period diffuser architecture that progressively generates target series from long to short periods.

\begin{figure*}[t!]
	\centering
	\subfloat[Swin Attention]{\includegraphics[width=.43\linewidth]{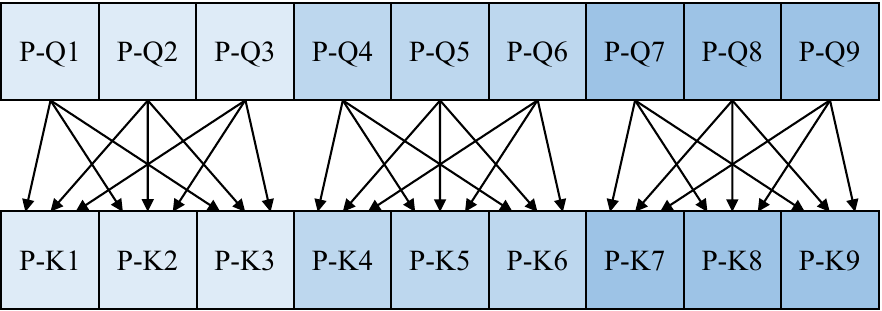}\label{swin}}\hspace{3mm}
	\subfloat[LogSparse Attention]{\includegraphics[width=.43\linewidth]{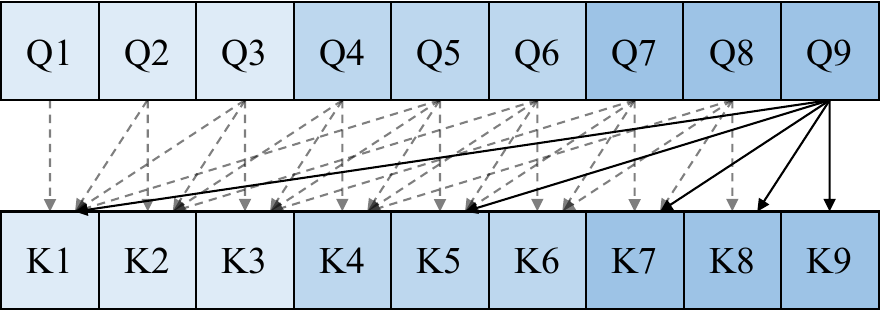}\label{lsa}}\hspace{3mm}
	\subfloat[ProbSparse Attention]{\includegraphics[width=.43\linewidth]{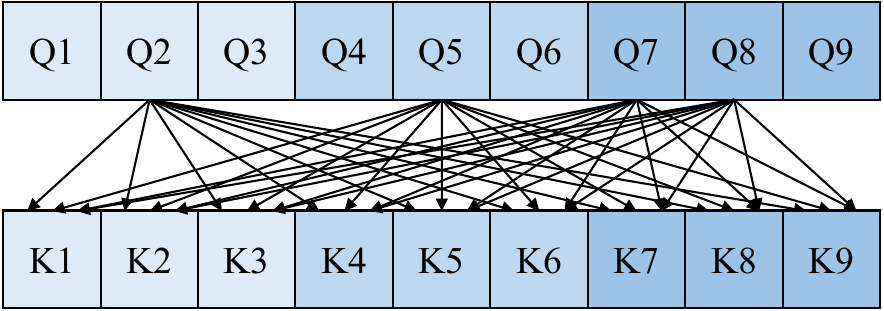}\label{psa}}\hspace{3mm}
	\subfloat[Auto-Correlation]{\includegraphics[width=.43\linewidth]{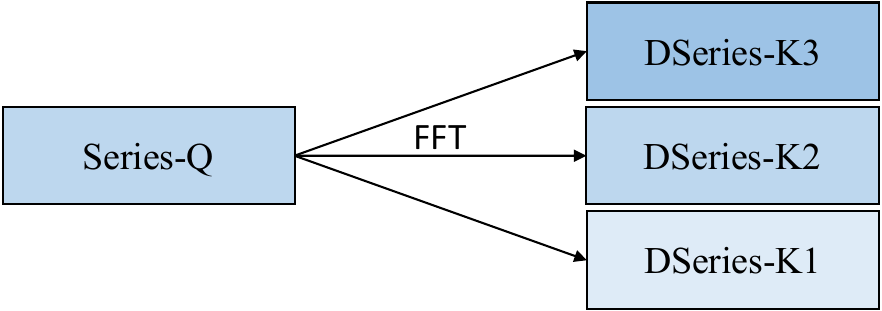}\label{ac}}\hspace{3mm}
	\subfloat[Segment-Correlation]{\includegraphics[width=.43\linewidth]{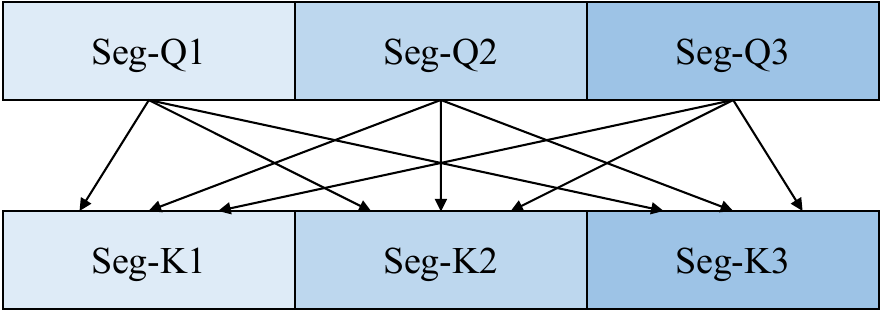}\label{sc}}\hspace{3mm}
	\subfloat[Isometric Convolution]{\includegraphics[width=.43\linewidth]{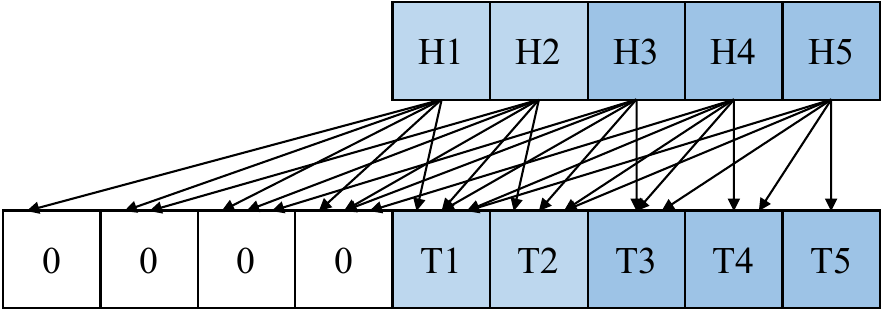}\label{ic}}\hspace{3mm}
	\subfloat[Period Attention]{\includegraphics[width=.43\linewidth]{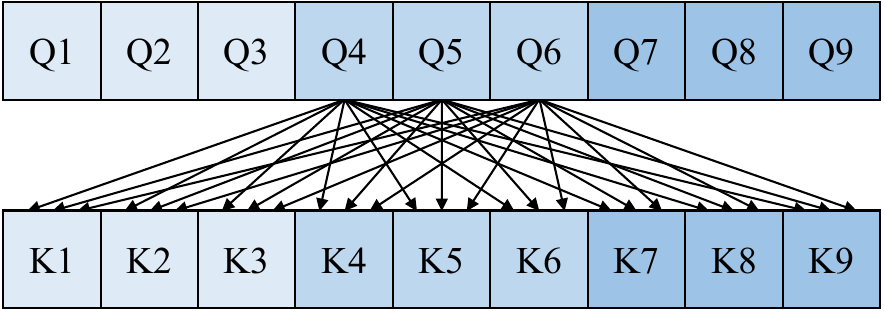}\label{pa}}\hspace{3mm}
	\subfloat[Sparse Period Attention]{\includegraphics[width=.43\linewidth]{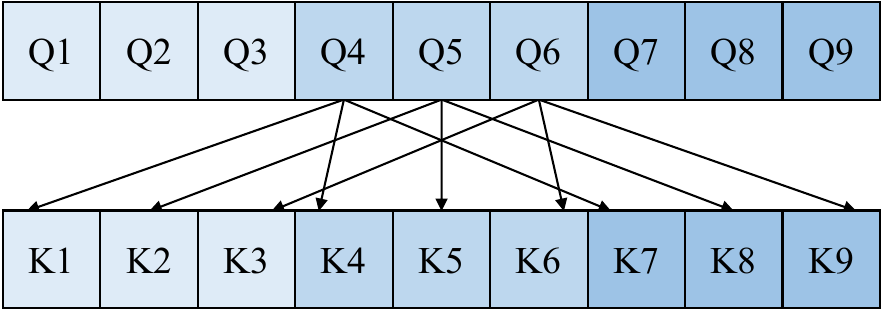}\label{spa}}
	\caption{(a) Local attention mechanisms, like Swin Transformer, query the similarity of patches in fixed windows. They require either deep layers or additional shifted window layers to realize cross-window aggregation. (b) LogSparse attention mechanism queries a fixed subset of keys to reduce complexity. (c) ProbSparse attention mechanism focuses on top-u queries and all keys. (d) Auto-Correlation in Autoformer uses FFT to find the correlation between the query series and key series with different time delays. (e) Segment-Correlation mechanism in Preformer adopts dot product of matrices to compare the similarity of two segments in time series. (f) Isometric convolution uses large kernels after padding to achieve a global temporal inductive bias. (g) PAM naturally exchanges the semantic information between adjacent periods. (h) SPAM reduces more redundant connections than PAM.}
	\label{attns}
\end{figure*}

\subsection{Period Attention and Sparse Period Attention}\label{3.1}
Time series data are characterized by their continuous nature, exhibiting strong locality. Recent studies \citep{nieTimeSeriesWorth2023, yecheNeighborhoodContrastiveLearning, tonekaboniUNSUPERVISEDREPRESENTATIONLEARNING2021} have demonstrated the significance of incorporating locality awareness in time series analysis. 
Fig. (\ref{attns}) exhibits the differences between well-known token mixers and the proposed mechanisms where Q, K, H, and T represent point-wise features in time series. P-Q, Seg-Q, and Series-Q are patch-wise, segment-wise, and series-wise features, respectively. Existing variants of Transformer models either require deeper layer stacking for global pattern analysis, like Swin Transformer \citep{liuSwinTransformerHierarchical2021} in Fig. (\ref{swin}) and LogSparse Transformer in Fig. (\ref{lsa}), or overlook local features, such as Informer depicted in Fig. (\ref{psa}) and Autoformer in Fig. (\ref{ac}). Preformer in Fig. (\ref{sc}) achieves global-local awareness through segment-wise attention, but it is not able to process in-segment details. Feature extraction of convolution-based models, like isometric convolution in Fig. (\ref{ic}), is limited by the sharing weights. These models fail to fully explore the relationships among periods of various lengths. To address these challenges and maximize the performance of the attention mechanism, we propose the period attention mechanism (PAM) illustrated in Fig. (\ref{pa}). Our approach begins with extracting short periods of length $P$ to mitigate the influence of composite patterns of shorter periods when processing longer periods. Let $\bm{X} \in \mathbb{R}^{L\times C}$ be an MTS, which consists of $C$ variables and has a length of $L$. Let $\bm{X}_{t:t+P-1} \in \mathbb{R}^{P \times C}$ denote the MTS data from time step $t$ to $t+P - 1$. Each univariate series from $\bm{X}$ is initially embedded into $\bm{Z}\in \mathbb{R}^{L\times D}$ as the input of PeriodNet, where $D$ stands for the dimension of the input embedding. The TSF task is to generate future values from time step $L+1$ to $L+T$, denoted by $\bm{X}_{L+1:L+T}$.

As one of the most important features in TSF, periodicity directly decides the details in the predicted time series. Time series typically comprises periods of various lengths. Due to the influence of trend and random distortion, characteristics may significantly differ for periods located far apart. To obtain robust temporal representations from input time series, we adopt a period-based perspective and focus on adjacent periods, assuming that adjacent periods exhibit similar characteristics, such as trend consistency. Meanwhile, we preserve point-wise calculations to retain the detailed information within each period. Denote the output of PAM by $\bm{Z}_{p} \in \mathbb{R}^{L\times D}$ for further processing and the proposed PAM is defined as:
\begin{equation}
	\label{period attention}
	\bm{Z}_{p} = \text{MHA}(\bm{Z}_{t:t+P-1},\bm{Z}_{N_t}, \bm{Z}_{N_t}),
\end{equation}
where $\text{MHA}(Q, K, V)$ denotes the multi-head dot-product attention mechanism. Let $\bm{Z}_{N_t} = \{\bm{Z}_{t-P:t-1},$ $\bm{Z}_{t:t+P - 1}, \bm{Z}_{t+P:t+2\times P}\}$ be the neighboring periods of $\bm{Z}_{t:t+P - 1}$. By using a one-layer PAM, periods with length lower than $P$ can be analyzed.

Time series data can be recorded at different granularity, like every minute, every hour or every day. When the granularity of input time series is excessively fine-grained for the forecasting-oriented trend and periods features, some unnecessary fluctuations and redundant information may easily mislead the convergence of neural networks, resulting in significantly performance degradation. To address the temporal redundancy in time series data, we introduce the sparse period attention mechanism (SPAM) depicted in Fig. (\ref{spa}) to properly reduce the resolution of the input time series and filter out some task-irrelevant details.
SPAM functions as an attention-based dilated convolution operation, eliminating redundant connections between queries and keys. Therefore, it can prevent the attention mechanism from learning unnecessary temporal patterns and reduce the risk of overfitting.

However, not all periods are valuable for TSF. Some frequencies may introduce noise and decrease prediction accuracy. Most existing local attention methods expand receptive fields by adding more layers, which fails to mitigate noise interference. An effective time-series encoder should actively locate significant periods for accurate predictions. Therefore, we introduce a period router after PAM/SPAM to dynamically identify the appropriate receptive fields. Let $\bm{\hat{Z}}_r \in \mathbb{R}^{r\times D}$ and $\bm{Z}_r \in  \mathbb{R}^{L\times D}$ be the routing features and output of the period router, respectively, as defined below:
\begin{equation}
	\label{route1}
	\bm{\hat{Z}}_r = \text{MHA}(\bm{M}, \bm{Z}_{p}, \bm{Z}_{p}),
\end{equation}
\begin{equation}
	\label{route2}
	\bm{Z}_r = \text{MHA}(\bm{Z}_{p}, \bm{\hat{Z}}_r, \bm{\hat{Z}}_r),
\end{equation}
where $\bm{M} \in \mathbb{R}^{r\times D}$ is a learnable matrix and $r$ is a constant representing the length of the period router. $\bm{M}$ queries similarity among the output of PAM/SPAM, $\bm{Z}_{p}$, to aggregate semantic information among all periods of length $P$ in hidden features and generate the routing features, $\bm{\hat{Z}}_r$. $\bm{Z}_{p}$ queries the global patterns within the routing features to form the output of the period router, $\bm{Z}_r$. 
The period router can dynamically extend receptive fields, identifying periods of much longer scales. This allows PAM/SPAM in the subsequent encoder block to extract significantly longer periods without the need for excessive layer stacking. Furthermore, each encoder block ends up with a feed-forward layer responsible for cross-channel feature aggregation. Let $\bm{Z}_n \in \mathbb{R}^{L\times D}$ be the output of the encoder block, as defined below:
\begin{equation}
	\label{ff}
	\bm{Z}_n = \text{Norm}(\bm{Z}_r + \text{FFN}(\bm{Z}_r)),
\end{equation}
where $\text{FFN}$ represents the feed forward layers for point-wise channel modeling and $\text{Norm}$ is layer normalization.

\subsection{Iterative Grouping Mechanism}
The multivariate joint modeling approach in Informer treats all time series as a single entity after the embedding stage, which may cause cross-variate interference during the prediction phase. The channel independence assumption in PatchTST ignores the cross-variable dependencies, resulting in limited prediction performance. The cross-variate modeling in Crossformer necessitates additional attention computations with a high complexity among all variables.

\begin{figure*}[t]
	\centering
	\includegraphics[width=1.\textwidth]{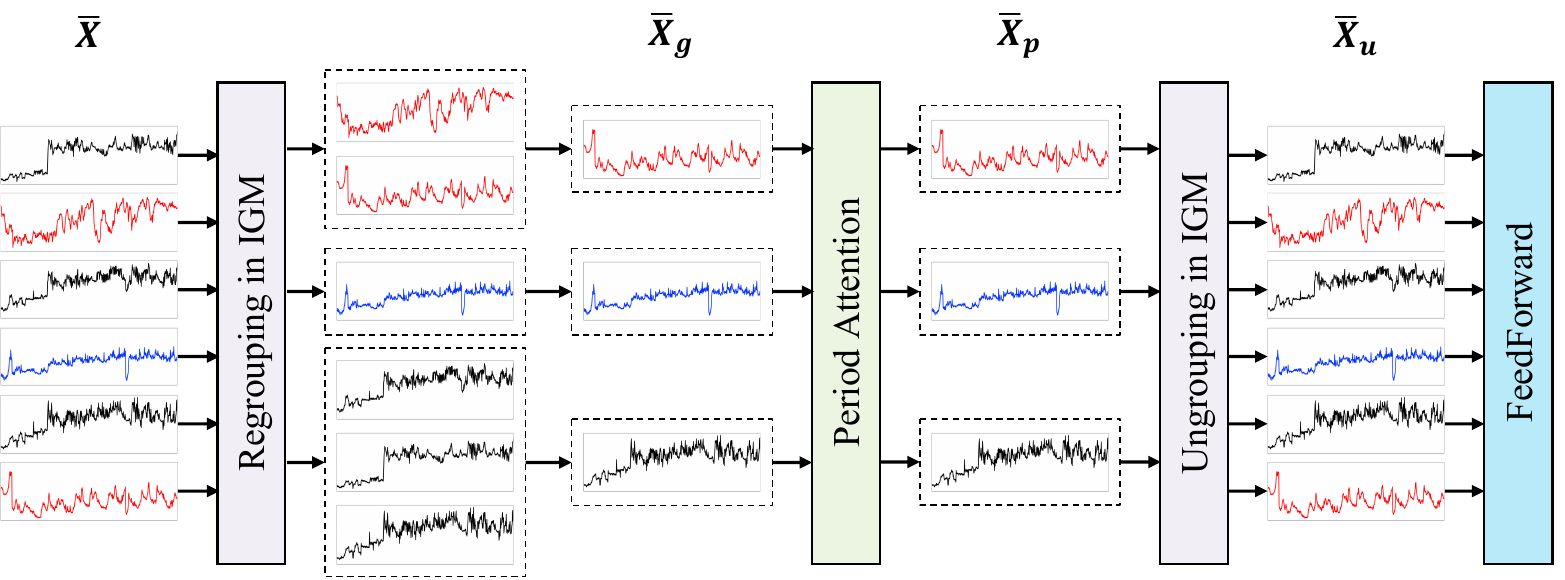} 
	\caption{In the iterative grouping mechanism, the regrouping process projects the input time series into $G$ groups, i.e. $G$ synthetic time series. The ungrouping process projects them back to the $C$-dim space.}
	\label{igm}
\end{figure*}

We make full use of the advantages of these methods above and introduce the iterative grouping mechanism (IGM). IGM primarily learns similar features across multiple variates, like those in the frequency domain, as demonstrated in Fig. (\ref{igm}). Different time series may exhibit similar features at periods of various lengths. For example, in the first encoder block, time series 1 and 3 can be grouped together for the analysis in the period of length 8. Besides, in the second encoder block, time series 1 and 2 can be grouped together for the analysis in the  period of length 27. To enable flexible grouping of time series based on the period length, IGM incorporates a regrouping stage before PAM and an ungrouping stage after PAM, as depicted in Fig. (\ref{igm}). The regrouping stage allows for the rearrangement of MTS, while the ungrouping stage restores the original arrangement right after PAM. Denote the input MTS comprising $C$ variables and $L$ time steps by $\bm{\bar{X}} \in \mathbb{R}^{L\times D \times C}$, and the regrouping stage is formulated as:
\begin{equation}
	\label{group}
	\bm{\bar{X}}_g = \sigma(\bm{\bar{X}}\bm{W}_{g_1}+\bm{b}_{g_1})\bm{W}_{g_2}+\bm{b}_{g_2},
\end{equation}
where $\bm{W}_{g_1} \in \mathbb{R}^{C\times h_g}$ and $\bm{W}_{g_2} \in \mathbb{R}^{h_g\times G}$ are learnable matrices to identify the similarity of different input time series while $\bm{b}_{g_1}$ and $\bm{b}_{g_2}$ are biases. $h_g$ represents the number of dimensions in the activation layer $\sigma$.
The output time series $\bm{\bar{X}}_g \in \mathbb{R}^{L\times D\times G}$ are composed of $G$ groups. Following the regrouping process, the grouped time series are individually processed by PAM and transformed into $\bm{\bar{X}}_p$.
The ungrouping stage aims to preserve the respective characteristics of each time series, preventing disturbances during the prediction caused by features collected from other time series. The ungrouping process is formulated as follows:
\begin{equation}
	\label{ungroup}
	\bm{\bar{X}}_u = \sigma(\bm{\bar{X}}_p\bm{W}_{u_1}+\bm{b}_{u_1})\bm{W}_{u_2} + \bm{b}_{u_2},
\end{equation}
where $\bm{W}_{u_1} \in \mathbb{R}^{G\times h_g}$ and $\bm{W}_{u_2} \in \mathbb{R}^{h_g\times C}$ are learnable ungrouping matrices while $\bm{b}_{u_1}$ and $\bm{b}_{u_2}$ are biases. $\bm{\bar{X}}_u$ is the ungrouped time series.
The output of the ungrouping process, $\bm{\bar{X}}_u$, is finally fed into the feed-forward layer for channel-wise aggregation for each variable.

The number of groups is directly influenced by the number of dimensions and the complexity of cross-variate dependencies. Typically, an increase in the number of dimensions in the input time series necessitates a larger number of groups. Similarly, in cases where cross-variate dependencies are complex, a large number of groups is also required.

\begin{figure}[t]
	\centering
	\includegraphics[width=.6\columnwidth]{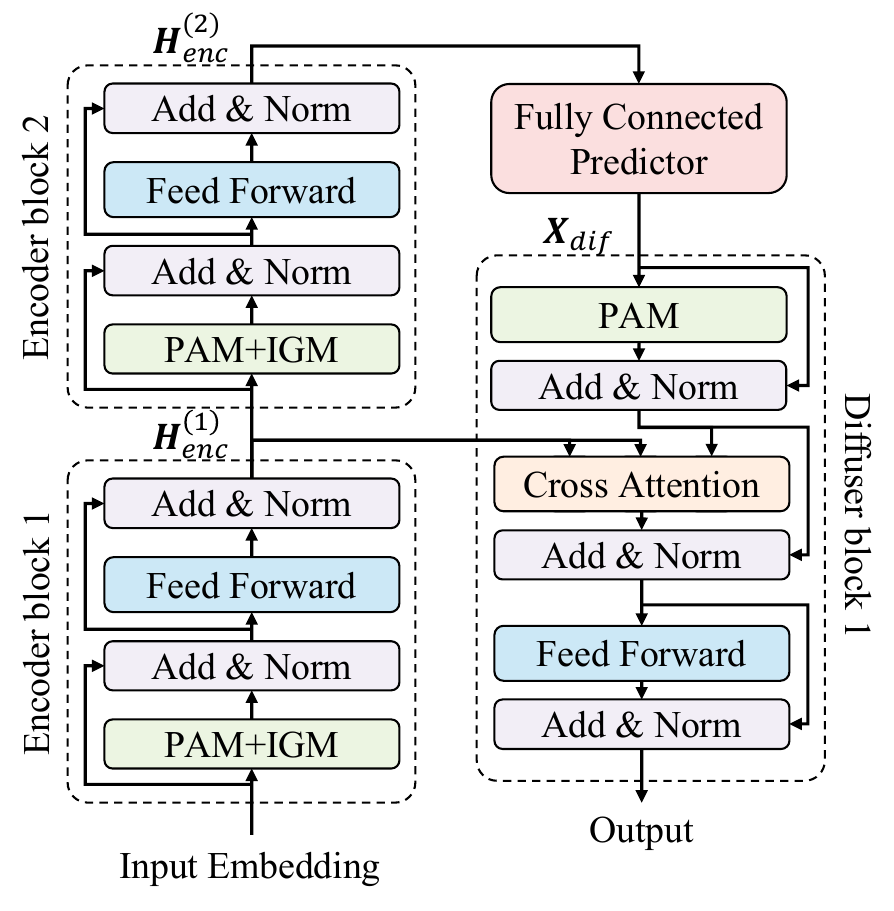} 
	\caption{Structure of PeriodNet.}
	\label{diffuser}
\end{figure}

\subsection{Period Diffuser}
The traditional predictor based on the autoregressive decoder suffers from error accumulation. Recently, most predictors have utilized generative style decoders and fully connected layers for single-step prediction based on the output features of the encoder. However, as discussed in DLinear \citep{zengAreTransformersEffective2022}, using a predefined positional embedding may not capture sufficient temporal relations in the target time series. Consequently, these predictors either fail to learn suitable temporal relationships or cannot accurately aggregate short- and long-period features from different encoder blocks within the target time series.

To fully leverage the multi-period features extracted by each encoder block, we revamp the architecture of the vanilla Transformer and introduce an encoder-diffuser architecture called period diffuser as shown in Fig. (\ref{diffuser}). PeriodNet comprises a PAM-based encoder, a fully connected predictor, and a period diffuser. The input MTS are jointly analyzed at the encoder side and the target MTS are separately predicted, thereby avoiding interference from multiple variables at the diffusion stage. The fully connected predictor and period diffuser use the same set of parameters when predicting the $C$ time series, reducing memory consumption.
Since the last encoder block naturally focuses on the longest periods in the given time series, we map the final output of the encoder, i.e. $\bm{H}_{enc}^{(2)}$ in Fig. (\ref{diffuser}), to the coarse-grained prediction of the target time series $\bm{X}_{dif}$. This approach significantly differs from other Transformer variants that employ elaborate embeddings as decoder inputs. Moreover, the fully connected predictor eliminates the need for a predefined positional embedding and facilitates the learning of adaptive temporal dependencies during training.

The period diffuser consists of $N_{enc}$ encoder blocks and $N_{dif}$ diffuser blocks. Let $\bm{X}_{dif}$ be the input to the period diffuser. Inspired by the diffusion process, the period diffuser employs cross attention to generate each univariate time series, processing from long periods to short periods gradually. This process is mathematically formulated as:
\begin{equation}
	\label{ca}
	\bm{Z}_c^{(i)} = \text{MHA}(\bm{Z}_{d}^{(i)}, \bm{H}_{enc}^{(N_{enc}-i)}, \bm{H}_{enc}^{(N_{enc}-i)}),
\end{equation}
where $\bm{Z}_{d}^{(i)}\in\mathbb{R}^{T\times D}$ in the $i$-th block of the period diffuser is the univariate time series output by PAM, $i = 1, ..., N_{dif}$. $\bm{H}_{enc}^{(N_{enc}-i)}$ stands for the output of the $(N_{enc}-i)$-th encoder block. $\bm{Z}_c^{(i)}\in\mathbb{R}^{T\times D}$ denotes the output of the cross attention process. The final predicted time series are generated through the feed-forward layer.

By leveraging the cross attention between each diffuser block and its counterpart encoder block, the target time series are generated in a progressive manner from long periods to short periods. This process results in better predictions compared to those obtained from traditional encoder-decoder architectures. Note that the period diffuser's inference speed remains unaffected by the length of prediction. PeriodNet is capable of achieving efficient and accurate predictions even in long-term TSF scenarios.

\section{Experiment}
\subsection{Experimental Setup}
\textbf{Datasets} Partial time series in four ETT (Electricity Transformer Temperature) datasets, namely ETTh1, ETTh2, ETTm1, and ETTm2, are selected for univariate TSF tests. Eight well-known datasets are selected for multivariate TSF tests, i.e. Weather, Electricity, ILI (Influenza-Like Illness), Exchange, and the four ETT datasets, to facilitate a comprehensive comparison. 
The four ETT\footnote{https://github.com/zhouhaoyi/ETDataset} datasets collect electricity time series data from seven indicators, including temperature and load. The record granularity in ETTh1 and ETTh2 is at an hourly level, whereas in ETTm1 and ETTm2, it is at a five-minute level.
ILI\footnote{https://gis.cdc.gov/grasp/fluview/fluportaldashboard.html} dataset contains medical time series data from seven variables. It captures the ratio of influenza-like illness patients to the total number of patients. The record granularity in this dataset is on a weekly basis.
Weather\footnote{https://www.bgc-jena.mpg.de/wetter/} dataset captures various weather indicators, such as temperature, pressure, humidity, and more. It encompasses time series generated from 21 variables, and the record granularity is at a ten-minute level.
Exchange\footnote{https://github.com/laiguokun/multivariate-time-series-data} dataset comprises economic time series data, specifically recording the daily fluctuation of exchange rates in eight countries.
Electricity\footnote{https://archive.ics.uci.edu/ml/datasets/ElectricityLoadDiagrams20112014} dataset records the electricity consumption of 321 customers. The record granularity in this dataset is at an hourly level.

The four ETT datasets are chronologically split by the ratio of 6:2:2 to generate the training, validation and test sets. All other datasets are split by the ratio of 7:1:2.

\noindent\textbf{Baselines} We refer to the proposed PAM-based model as PeriodNet and the SPAM-based model as SPeriodNet. Both of them are compared with six state-of-the-art models, namely PatchTST/64 \citep{nieTimeSeriesWorth2023}, DLinear \citep{zengAreTransformersEffective2022}, FEDformer \citep{zhouFEDformerFrequencyEnhanced2022}, Autoformer \citep{wuAutoformerDecompositionTransformers2021}, Informer \citep{zhouInformerEfficientTransformer2021a}, and LogSparse Transformer \citep{liEnhancingLocalityBreaking2020} (referred to as LogTrans). These models incorporate various underlying techniques, such as low-rank attention, vanilla attention, CNNs, and FCNs.

\begin{figure}[h]
	\centering
	\includegraphics[width=.6\linewidth]{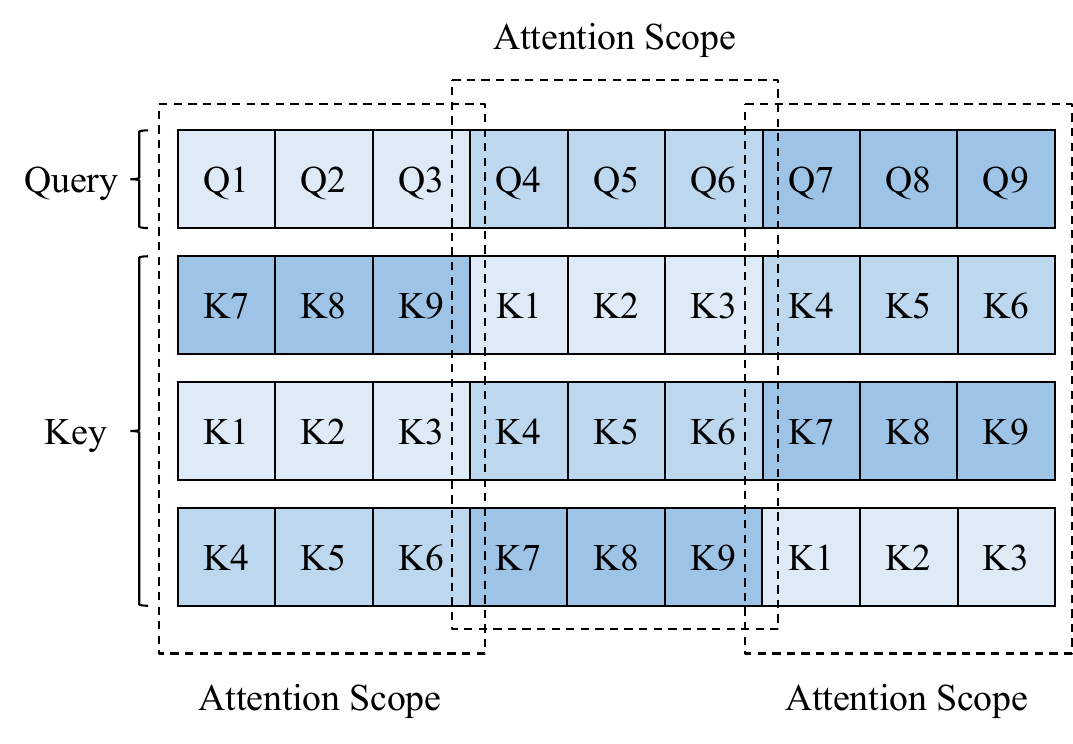}
	\caption{Implementation of PAM with a period of length 3.}
	\label{period_imp}
\end{figure}

\noindent\textbf{Implementation Details} PeriodNet and SPeriodNet are trained by the Adam optimizer with L2 loss. In almost all experiments, they consist of two encoder blocks and one diffuser block. The numbers of encoder and diffuser blocks may increase when handling complex TSF tasks, e.g. Electricity. In order to avoid noise interference, the selection of the period length $P$ shouldn't be too small, and we test it from 8 in all experiments. For the number of groups denoted by $G$ in IGM, the lower the similarity between different time series is, the larger the number of groups is usually required. In the experiments of eight datasets, $G$ ranges from 1 to 8. The prediction lengths are 24, 36, 48, and 60 for the ILI dataset, and 96, 192, 336, and 720 for all other datasets, respectively. The dimensions of the input embedding range from 8 to 128, and those of hidden features in the feed-forward layers range from 32 to 512.
We utilize MSE and MAE as performance metrics. MSE is sensitive to outliers while MAE provides an overall measure of the prediction model's performance. All experiments are implemented in Pytorch on four Nvidia Titan V 12G GPUs.

The implementation of the PAM is illustrated in Fig. (\ref{period_imp}). When dealing with a period of length 3, the key sequences are formulated by shifting the projected sequence by one period length in forward and backward directions. The computation of the sparse period attention mechanism adopts similar procedures after an additional dilated sampling process. 
This approach facilitates the transmission of semantic information across neighboring periods, thereby offering a strengthened capability for flexible modeling. This important capability achieves more efficient feature extraction compared with conventional local attention mechanisms, such as those in the Swin Transformer architecture.

\begin{table*}[t]
	\centering
	\caption{Multivariate long-term TSF results on 8 datasets. Note that the best and second best results are in bold and underlined, respectively.}
	\resizebox{1.0\linewidth}{!}{
		\begin{tabular}{@{}cc|cc|cc|cc|cc|cc|cc|cc|cc@{}}
			\toprule
			\multicolumn{2}{c|}{Methods}                                     & \multicolumn{2}{c|}{PeriodNet} & \multicolumn{2}{c|}{SPeriodNet} & \multicolumn{2}{c|}{PatchTST/64} & \multicolumn{2}{c|}{DLinear}    & \multicolumn{2}{c|}{FEDformer} & \multicolumn{2}{c|}{Autoformer} & \multicolumn{2}{c|}{Informer} & \multicolumn{2}{c}{LogTrans} \\ \midrule
			\multicolumn{2}{c|}{Metric}                                      & MSE             & MAE             & MSE              & MAE             & MSE             & MAE            & MSE            & MAE            & MSE            & MAE           & MSE            & MAE            & MSE           & MAE           & MSE           & MAE          \\ \midrule
			\multicolumn{1}{c|}{\multirow{4}{*}{\rotatebox{90}{Weather}}}     & 96           & \underline{0.146}     & \textbf{0.195}  & \textbf{0.145}   & \underline{0.197}     & 0.149           & 0.198          & 0.176          & 0.237          & 0.238          & 0.314         & 0.249          & 0.329          & 0.354         & 0.405         & 0.458         & 0.490        \\
			\multicolumn{1}{c|}{}                             & 192          & \underline{0.192}     & 0.242           & \textbf{0.189}   & \textbf{0.240}  & 0.194           & \underline{0.241}    & 0.220          & 0.282          & 0.275          & 0.329         & 0.325          & 0.370          & 0.419         & 0.434         & 0.658         & 0.589        \\
			\multicolumn{1}{c|}{}                             & \textit{336} & \underline{0.243}     & \underline{0.282}     & \textbf{0.242}   & \textbf{0.278}  & 0.245           & \underline{0.282}    & 0.265          & 0.319          & 0.339          & 0.377         & 0.351          & 0.391          & 0.583         & 0.543         & 0.797         & 0.652        \\
			\multicolumn{1}{c|}{}                             & \textit{720} & \underline{0.315}     & \textbf{0.329}  & 0.317            & \underline{0.330}     & \textbf{0.314}  & 0.334          & 0.323          & 0.362          & 0.389          & 0.409         & 0.415          & 0.426          & 0.916         & 0.705         & 0.869         & 0.675        \\ \midrule
			\multicolumn{1}{c|}{\multirow{4}{*}{\rotatebox{90}{Electricity}}} & 96           & \textbf{0.128}  & \underline{0.223}     & \underline{0.129}      & \underline{0.223}     & \underline{0.129}     & \textbf{0.222} & 0.140          & 0.237          & 0.186          & 0.302         & 0.196          & 0.313          & 0.304         & 0.393         & 0.258         & 0.357        \\
			\multicolumn{1}{c|}{}                             & 192          & \textbf{0.147}  & \textbf{0.239}  & \textbf{0.147}   & 0.241           & \textbf{0.147}  & \underline{0.240}    & 0.153          & 0.249          & 0.197          & 0.311         & 0.211          & 0.324          & 0.327         & 0.417         & 0.266         & 0.368        \\
			\multicolumn{1}{c|}{}                             & \textit{336} & \textbf{0.161}  & \textbf{0.258}  & \underline{0.163}      & 0.263           & \underline{0.163}     & \underline{0.259}    & 0.169          & 0.267          & 0.213          & 0.328         & 0.214          & 0.327          & 0.333         & 0.422         & 0.280         & 0.380        \\
			\multicolumn{1}{c|}{}                             & \textit{720} & \underline{0.191}     & \textbf{0.283}  & \textbf{0.188}   & \underline{0.286}     & 0.197           & 0.290          & 0.203          & 0.301          & 0.233          & 0.344         & 0.236          & 0.342          & 0.351         & 0.427         & 0.283         & 0.376        \\ \midrule
			\multicolumn{1}{c|}{\multirow{4}{*}{\rotatebox{90}{ILI}}}         & 24           & \underline{1.387}     & \textbf{0.746}  & 1.451            & 0.789           & \textbf{1.319}  & \underline{0.754}    & 2.215          & 1.081          & 2.624          & 1.095         & 2.906          & 1.182          & 4.657         & 1.499         & 4.480         & 1.444        \\
			\multicolumn{1}{c|}{}                             & 36           & \underline{1.495}     & \underline{0.811}     & \textbf{1.399}   & \textbf{0.783}  & 1.579           & 0.870          & 1.963          & 0.963          & 2.516          & 1.021         & 2.585          & 1.038          & 4.650         & 1.463         & 4.799         & 1.467        \\
			\multicolumn{1}{c|}{}                             & \textit{48}  & \underline{1.429}     & \underline{0.756}     & \textbf{1.259}   & \textbf{0.730}  & 1.553           & 0.815          & 2.130          & 1.024          & 2.505          & 1.041         & 3.024          & 1.145          & 5.004         & 1.542         & 4.800         & 1.468        \\
			\multicolumn{1}{c|}{}                             & \textit{60}  & \underline{1.454}     & 0.795           & \textbf{1.432}   & \textbf{0.768}  & 1.470           & \underline{0.788}    & 2.368          & 1.096          & 2.742          & 1.122         & 2.761          & 1.114          & 5.071         & 1.543         & 5.278         & 1.560        \\ \midrule
			\multicolumn{1}{c|}{\multirow{4}{*}{\rotatebox{90}{Exchange}}}    & 96           & \underline{0.080}     & \underline{0.198}     & \textbf{0.079}   & \textbf{0.196}  & 0.096           & 0.217          & 0.081          & 0.203          & 0.148          & 0.278         & 0.197          & 0.323          & 0.847         & 0.752         & 0.968         & 0.812        \\
			\multicolumn{1}{c|}{}                             & 192          & \underline{0.164}     & \underline{0.293}     & \underline{0.164}      & \textbf{0.291}  & 0.201           & 0.322          & \textbf{0.157} & \underline{0.293}    & 0.271          & 0.380         & 0.300          & 0.369          & 1.204         & 0.895         & 1.040         & 0.851        \\
			\multicolumn{1}{c|}{}                             & \textit{336} & \textbf{0.280}  & \textbf{0.385}  & \underline{0.283}      & \underline{0.387}     & 0.373           & 0.448          & 0.305          & 0.414          & 0.460          & 0.500         & 0.509          & 0.524          & 1.672         & 1.036         & 1.659         & 1.081        \\
			\multicolumn{1}{c|}{}                             & \textit{720} & \textbf{0.607}  & \textbf{0.582}  & 0.683            & 0.625           & 0.913           & 0.704          & \underline{0.643}    & \underline{0.601}    & 1.195          & 0.841         & 1.447          & 0.941          & 2.478         & 1.310         & 1.941         & 1.127        \\ \midrule
			\multicolumn{1}{c|}{\multirow{4}{*}{\rotatebox{90}{ETTh1}}}       & 96           & \underline{0.368}     & \underline{0.393}     & \textbf{0.364}   & \textbf{0.391}  & 0.370           & 0.400          & 0.375          & 0.399          & 0.376          & 0.415         & 0.435          & 0.446          & 0.941         & 0.769         & 0.878         & 0.740        \\
			\multicolumn{1}{c|}{}                             & 192          & \textbf{0.405}  & 0.417           & 0.409            & \textbf{0.416}  & 0.413           & 0.429          & \textbf{0.405} & \textbf{0.416} & 0.423          & 0.446         & 0.456          & 0.457          & 1.007         & 0.786         & 1.037         & 0.824        \\
			\multicolumn{1}{c|}{}                             & \textit{336} & \textbf{0.420}  & \underline{0.429}     & \underline{0.422}      & \textbf{0.425}  & \underline{0.422}     & 0.440          & 0.439          & 0.443          & 0.444          & 0.462         & 0.486          & 0.487          & 1.038         & 0.784         & 1.238         & 0.932        \\
			\multicolumn{1}{c|}{}                             & \textit{720} & \textbf{0.427}  & \textbf{0.451}  & \underline{0.440}      & \underline{0.453}     & 0.447           & 0.468          & 0.472          & 0.490          & 0.469          & 0.492         & 0.515          & 0.517          & 1.144         & 0.857         & 1.135         & 0.852        \\ \midrule
			\multicolumn{1}{c|}{\multirow{4}{*}{\rotatebox{90}{ETTh2}}}       & 96           & \underline{0.278}     & \textbf{0.334}  & 0.279            & 0.341           & \textbf{0.274}  & \underline{0.337}    & 0.289          & 0.353          & 0.332          & 0.374         & 0.332          & 0.368          & 1.549         & 0.952         & 2.116         & 1.197        \\
			\multicolumn{1}{c|}{}                             & 192          & \underline{0.344}     & \textbf{0.382}  & 0.347            & 0.384           & \textbf{0.341}  & \textbf{0.382} & 0.383          & 0.418          & 0.407          & 0.446         & 0.426          & 0.434          & 3.792         & 1.542         & 4.315         & 1.635        \\
			\multicolumn{1}{c|}{}                             & \textit{336} & \underline{0.346}     & \underline{0.392}     & 0.351            & 0.393           & \textbf{0.329}  & \textbf{0.384} & 0.448          & 0.465          & 0.400          & 0.447         & 0.477          & 0.479          & 4.215         & 1.642         & 1.124         & 1.604        \\
			\multicolumn{1}{c|}{}                             & \textit{720} & \textbf{0.374}  & \textbf{0.420}  & 0.427            & 0.448           & \underline{0.379}     & \underline{0.422}    & 0.605          & 0.551          & 0.412          & 0.469         & 0.453          & 0.490          & 3.656         & 1.619         & 3.188         & 1.540        \\ \midrule
			\multicolumn{1}{c|}{\multirow{4}{*}{\rotatebox{90}{ETTm1}}}       & 96           & \underline{0.284}     & \textbf{0.337}  & 0.290            & \underline{0.339}     & 0.293           & 0.346          & \textbf{0.229} & 0.343          & 0.326          & 0.390         & 0.510          & 0.492          & 0.626         & 0.560         & 0.600         & 0.546        \\
			\multicolumn{1}{c|}{}                             & 192          & \underline{0.323}     & \textbf{0.360}  & \textbf{0.321}   & \underline{0.362}     & 0.333           & 0.370          & 0.335          & 0.365          & 0.365          & 0.415         & 0.514          & 0.495          & 0.725         & 0.619         & 0.837         & 0.700        \\
			\multicolumn{1}{c|}{}                             & \textit{336} & \textbf{0.348}  & \textbf{0.381}  & \underline{0.353}      & \textbf{0.381}  & 0.369           & 0.392          & 0.369          & 0.386          & 0.392          & 0.425         & 0.510          & 0.492          & 1.005         & 0.741         & 1.124         & 0.832        \\
			\multicolumn{1}{c|}{}                             & \textit{720} & \textbf{0.396}  & \textbf{0.411}  & \underline{0.405}      & \textbf{0.411}  & 0.416           & 0.420          & 0.425          & 0.421          & 0.446          & 0.458         & 0.527          & 0.493          & 1.133         & 0.845         & 1.153         & 0.820        \\ \midrule
			\multicolumn{1}{c|}{\multirow{4}{*}{\rotatebox{90}{ETTm2}}}       & 96           & \underline{0.163}     & \textbf{0.250}  & \textbf{0.161}   & \textbf{0.250}  & 0.166           & 0.256          & 0.167          & 0.260          & 0.180          & 0.271         & 0.205          & 0.293          & 0.355         & 0.462         & 0.768         & 0.642        \\
			\multicolumn{1}{c|}{}                             & 192          & \underline{0.220}     & \underline{0.294}     & \textbf{0.213}   & \textbf{0.291}  & 0.223           & 0.296          & 0.224          & 0.303          & 0.252          & 0.318         & 0.278          & 0.336          & 0.595         & 0.586         & 0.989         & 0.757        \\
			\multicolumn{1}{c|}{}                             & \textit{336} & \underline{0.271}     & \textbf{0.326}  & \textbf{0.268}   & \underline{0.329}     & 0.274           & \underline{0.329}    & 0.281          & 0.342          & 0.324          & 0.364         & 0.343          & 0.379          & 1.270         & 0.871         & 1.334         & 0.872        \\
			\multicolumn{1}{c|}{}                             & \textit{720} & \underline{0.349}     & \textbf{0.377}  & \textbf{0.339}   & \underline{0.378}     & 0.362           & 0.385          & 0.397          & 0.421          & 0.410          & 0.420         & 0.414          & 0.419          & 3.001         & 1.267         & 3.048         & 1.328        \\ \bottomrule
		\end{tabular}
	}
	
	\label{mtsf}
\end{table*}

\subsection{Experimental Results and Analysis}
Multivariate TSF requires comprehensive ability in univariate time series analysis, cross-variate modeling, and robust prediction process. As shown in Table \ref{mtsf}, PeriodNet and SPeriodNet achieve the top two results in most cases. Specifically, PeriodNet obtains the lowest MSE in 29 cases and the lowest MAE in 31 cases, while SPeriodNet achieves the lowest MSE in 26 cases and the lowest MAE in 27 cases. Notably, when compared to models based on the vanilla encoder-decoder Transformer with a prediction length of 720, PeriodNet achieves 19\% MSE reduction on the weather dataset, 18\% on the electricity dataset, 47\% on the ILI dataset, and 49\% on the exchange dataset. SPeriodNet achieves 19\% MSE reduction on the weather dataset, 19\% on the electricity dataset, 48\% on the ILI dataset, and 43\% on the exchange dataset. These results demonstrate significance of the proposed period diffuser architecture.

It is worth noting that the prediction loss on the ILI dataset does not increase with an increase of the prediction length. We attribute this behavior to the lack of clear periodicity in shorter prediction lengths, which influences the period analysis in the encoder and hence leads to unstable prediction accuracy in the predictor.

\begin{figure}[h!]
	\centering
	\subfloat[Different input lengths.]{\includegraphics[width=.45\linewidth]{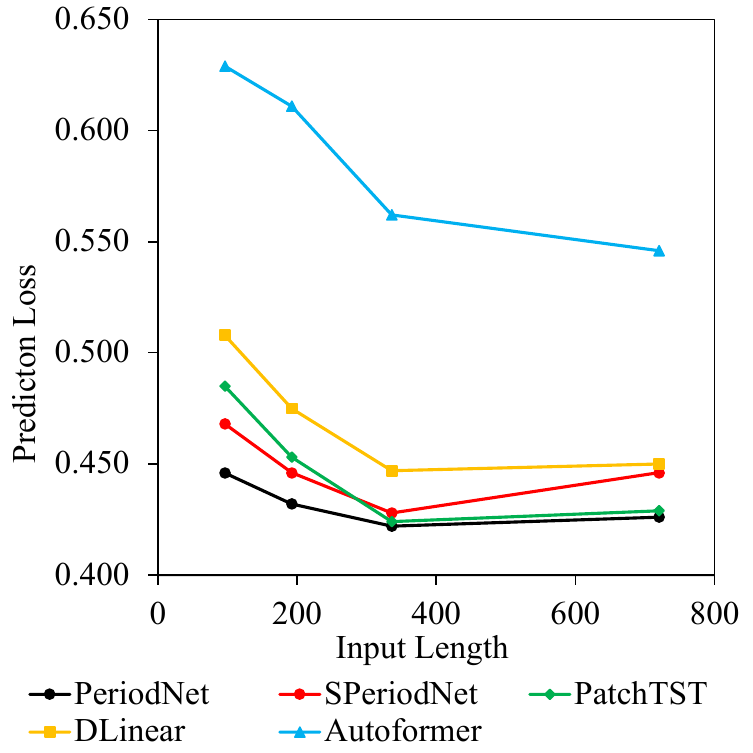}\label{diffinp}}
	\subfloat[Different prediction lengths.]{\includegraphics[width=.45\linewidth]{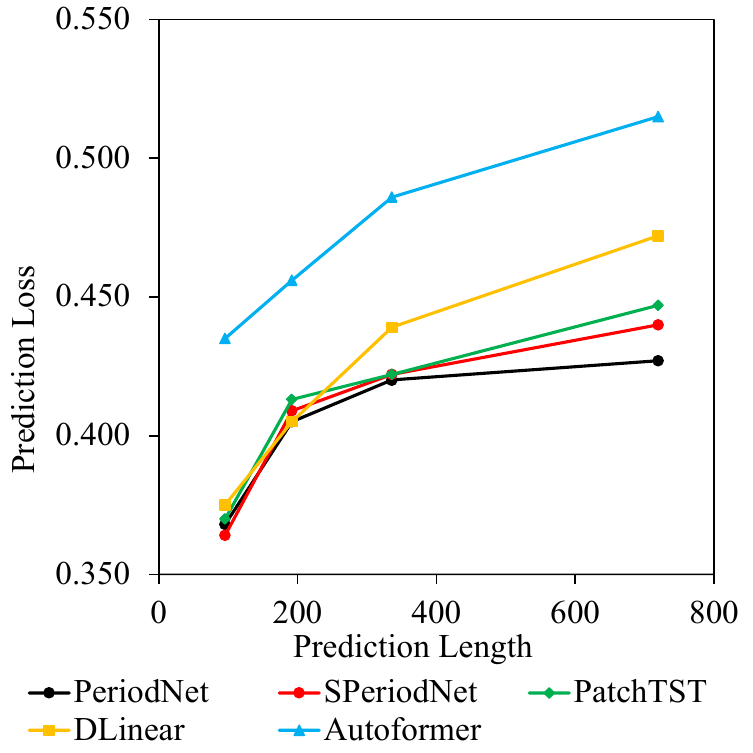}\label{diffpred}}
	\caption{Prediction loss results of five models on univariate time series from ETTh1 dataset.}
	\label{diff}
\end{figure}

To evaluate the overall prediction performance of PeriodNet and SPeriodNet, we vary the prediction length and input length on the univariate time series of the ETTh1 dataset, as illustrated in Fig. (\ref{diff}). As the input length varies in Fig. (\ref{diffinp}), PeriodNet consistently exhibits lower prediction errors compared to three other advanced models based on different techniques, highlighting the advantages of attention-based period analysis. Furthermore, according to Fig. (\ref{diffpred}), the proposed models gain more stable prediction performance compared to the other models as the prediction length increases.

In the context of univariate TSF, excellent abilities for temporal analysis, like global-local dependency mining and redundancy resistance, are usually demanded by token mixers. The effectiveness of a token mixer becomes more evident without considering MTS modeling. Besides, predictor is required to make full use of the features extracted from the associated encoder. As shown in Table \ref{utsf}, PeriodNet and SPeriodNet can effectively handle time series with various characteristics and achieve the top two results in most experiments. 

\begin{table*}[t]
	\centering
	\caption{Univariate long-term time series forecasting results on 4 datasets.}
	\resizebox{1.\linewidth}{!}{
		\begin{tabular}{@{}cc|cc|cc|cc|cc|cc|cc|cc|cc@{}}
			\toprule
			\multicolumn{2}{c|}{Methods}                                        & \multicolumn{2}{c|}{PeriodNet} & \multicolumn{2}{c|}{SPeriodNet} & \multicolumn{2}{c|}{PatchTST/64} & \multicolumn{2}{c|}{DLinear}    & \multicolumn{2}{c|}{FEDformer} & \multicolumn{2}{c|}{Autoformer} & \multicolumn{2}{c|}{Informer} & \multicolumn{2}{c}{LogTrans} \\ \midrule
			\multicolumn{2}{c|}{Metric}                                         & MSE             & MAE             & MSE              & MAE             & MSE             & MAE            & MSE            & MAE            & MSE          & MAE             & MSE            & MAE            & MSE           & MAE           & MSE            & MAE         \\ \midrule
			\multicolumn{1}{c|}{\multirow{4}{*}{\textit{\rotatebox{90}{ETTh1}}}} & 96           & \textbf{0.054}  & \textbf{0.178}  & \textbf{0.054}   & 0.181           & 0.059           & 0.189          & 0.056          & \underline{0.180}    & 0.079        & 0.215           & 0.071          & 0.206          & 0.193         & 0.377         & 0.283          & 0.468       \\
			\multicolumn{1}{c|}{}                                & 192          & \textbf{0.065}  & \underline{0.202}     & \underline{0.068}      & \textbf{0.201}  & 0.074           & 0.215          & 0.071          & 0.204          & 0.104        & 0.245           & 0.114          & 0.262          & 0.217         & 0.395         & 0.234          & 0.409       \\
			\multicolumn{1}{c|}{}                                & \textit{336} & \textbf{0.075}  & \textbf{0.217}  & \underline{0.076}      & \underline{0.219}     & \underline{0.076}     & 0.220          & 0.098          & 0.244          & 0.119        & 0.270           & 0.107          & 0.258          & 0.202         & 0.381         & 0.386          & 0.546       \\
			\multicolumn{1}{c|}{}                                & \textit{720} & \textbf{0.082}  & \underline{0.227}     & \textbf{0.082}   & \textbf{0.226}  & 0.087           & 0.236          & 0.189          & 0.359          & 0.142        & 0.299           & 0.126          & 0.283          & 0.183         & 0.355         & 0.475          & 0.629       \\ \midrule
			\multicolumn{1}{c|}{\multirow{4}{*}{\textit{\rotatebox{90}{ETTh2}}}} & 96           & \underline{0.128}     & \textbf{0.271}  & \textbf{0.127}   & 0.278           & 0.131           & 0.284          & 0.131          & 0.279          & \underline{0.128}  & \textbf{0.271}  & 0.153          & 0.306          & 0.213         & 0.373         & 0.217          & 0.379       \\
			\multicolumn{1}{c|}{}                                & 192          & \underline{0.150}     & \underline{0.306}     & \textbf{0.145}   & \textbf{0.303}  & 0.171           & 0.329          & 0.176          & 0.329          & 0.185        & 0.330           & 0.204          & 0.351          & 0.227         & 0.387         & 0.281          & 0.429       \\
			\multicolumn{1}{c|}{}                                & \textit{336} & \underline{0.169}     & \underline{0.330}     & \textbf{0.159}   & \textbf{0.323}  & 0.171           & 0.336          & 0.209          & 0.367          & 0.231        & 0.378           & 0.246          & 0.389          & 0.242         & 0.401         & 0.293          & 0.437       \\
			\multicolumn{1}{c|}{}                                & \textit{720} & \underline{0.217}     & \underline{0.372}     & \textbf{0.189}   & \textbf{0.351}  & 0.223           & 0.380          & 0.276          & 0.426          & 0.278        & 0.420           & 0.268          & 0.409          & 0.291         & 0.439         & 0.218          & 0.387       \\ \midrule
			\multicolumn{1}{c|}{\multirow{4}{*}{\textit{\rotatebox{90}{ETTm1}}}} & 96           & \underline{0.027}     & 0.124           & \underline{0.027}      & \textbf{0.123}  & \textbf{0.026}  & \textbf{0.123} & 0.028          & \textbf{0.123} & 0.033        & 0.140           & 0.056          & 0.183          & 0.109         & 0.277         & 0.049          & 0.171       \\
			\multicolumn{1}{c|}{}                                & 192          & \textbf{0.038}  & \textbf{0.150}  & \textbf{0.038}   & \underline{0.151}     & 0.040           & \underline{0.151}    & 0.045          & 0.156          & 0.058        & 0.186           & 0.081          & 0.216          & 0.151         & 0.310         & 0.157          & 0.317       \\
			\multicolumn{1}{c|}{}                                & \textit{336} & \underline{0.053}     & \underline{0.173}     & \textbf{0.051}   & \textbf{0.172}  & \underline{0.053}     & 0.174          & 0.061          & 0.182          & 0.084        & 0.231           & 0.076          & 0.218          & 0.427         & 0.591         & 0.289          & 0.459       \\
			\multicolumn{1}{c|}{}                                & \textit{720} & \textbf{0.070}  & \underline{0.203}     & \textbf{0.070}   & \textbf{0.202}  & 0.073           & 0.206          & 0.080          & 0.210          & 0.102        & 0.250           & 0.110          & 0.267          & 0.438         & 0.586         & 0.430          & 0.579       \\ \midrule
			\multicolumn{1}{c|}{\multirow{4}{*}{\textit{\rotatebox{90}{ETTm2}}}} & 96           & \textbf{0.062}  & \textbf{0.182}  & \underline{0.063}      & \underline{0.183}     & 0.065           & 0.187          & \underline{0.063}    & \underline{0.183}    & 0.067        & 0.198           & 0.065          & 0.189          & 0.088         & 0.225         & 0.075          & 0.208       \\
			\multicolumn{1}{c|}{}                                & 192          & \underline{0.093}     & \underline{0.230}     & \underline{0.093}      & \underline{0.230}     & \underline{0.093}     & 0.231          & \textbf{0.092} & \textbf{0.227} & 0.102        & 0.245           & 0.118          & 0.256          & 0.132         & 0.283         & 0.129          & 0.275       \\
			\multicolumn{1}{c|}{}                                & \textit{336} & \textbf{0.119}  & \underline{0.264}     & \underline{0.120}      & \underline{0.265}     & 0.121           & 0.266          & \textbf{0.119} & \textbf{0.261} & 0.130        & 0.279           & 0.154          & 0.305          & 0.180         & 0.336         & 0.154          & 0.302       \\
			\multicolumn{1}{c|}{}                                & \textit{720} & 0.173           & 0.324           & \underline{0.172}      & 0.324           & \underline{0.172}     & 0.322          & 0.175          & \textbf{0.320} & 0.178        & 0.325           & 0.182          & 0.335          & 0.300         & 0.435         & \textbf{0.160} & \underline{0.321} \\ \bottomrule
		\end{tabular}
	}
	\label{utsf}
\end{table*}

PAM and SPAM are essentially based on point-wise computation with a specific focus on similarity comparisons among adjacent periods. Therefore, they can aggregate the advantages of both patch-wise and point-wise algorithms. Different from patch-wise algorithms, PAM and SPAM can sufficiently mine multi-scale detailed information within patches. In contrast to point-wise algorithms, they can effectively leverage local information, encompassing comprehensive information from one time step to another, thereby avoiding overfitting noise interference irrelevant to forecasting.
We consider that PeriodNet excels when dealing with the relatively large record granularity and low temporal redundancy. Conversely, SPeriodNet is more likely to achieve superior performance when record granularity is small and temporal redundancy is high. Its sparse sampling method enables it to neglect many forecasting-irrelevant fluctuations, to some extent.

PatchTST and DLinear are close to each other in performance. As a patch-wise algorithm, PatchTST analyzes time series based on fixed-size patches to exploit local information and demonstrates superior performance compared with traditional point-wise algorithms, like Informer, and LogTrans. This underscores the assertion in \citep{duPreformerPredictiveTransformer} that data at individual time steps lack meaning. However, the choice of patch size directly influences the prediction accuracy of time series and determining a suitable patch size for a given dataset poses a non-trivial challenge. Consequently, we can attribute PatchTST's inferior performance compared with PeriodNet and SPeriodNet to its inability to capture detailed information within each patch.
DLinear employs point-wise operations after the decomposition process, enabling the differentiation of seasonal and trend components while extracting short-scale features. However, DLinear cannot effectively address the locality and temporal redundancy in time series, which makes it hard to identify global dependencies and easy to overfit some unnecessary details.
FEDformer and Autoformer, founded on frequency operations and series-wise fusion, exhibit better performance in mitigating noise interference compared with vanilla Transformer-based algorithms, i.e., Informer and LogTrans.

\subsection{Ablation Study}
To explicitly evaluate the effectiveness of each component in PeriodNet, we maintain the same network structures with the same random seed and perform additional ablation experiments on the ETT datasets. 

\subsubsection{Effect of period attention mechanism}
In this study, our experiments focus on univariate time series data from the ETTh1 dataset to analyze the effects of different attention-based token mixers, eliminating any irrelevant influences from MTS modeling. Each model uses the same predictor, namely the period diffuser, while employing distinct attention mechanisms. The ProbSparse Attention in Informer is adopted in the low-rank attention mechanism (LRAM), whereas the shifted window attention in Swin Transformer is used in the local attention mechanism (LAM). Each encoder is composed of two blocks, and each predictor consists of one diffuser block.

\begin{table}[h]
	\centering
	\caption{Performance comparison of different attention mechanisms with MSE on the ETTh1 dataset.  Note that Inp Len and Pred Len denote input length and prediction length, respectively.}
	\scriptsize
	\begin{tabular}{@{}c|cccc|cccc@{}}
		\toprule
		Inp Len     & \multicolumn{4}{c|}{96}                                           & \multicolumn{4}{c}{336}                                           \\ \midrule
		Pred Len & 96             & 192            & 336            & 720            & 96             & 192            & 336            & 720            \\ \midrule
		LRAM        & 0.056          & 0.072          & 0.083          & 0.087          & 0.058          & 0.075          & 0.084          & 0.087          \\
		FAM         & 0.058          & 0.075          & 0.086          & 0.091          & 0.061          & 0.076          & 0.080           & 0.090           \\
		LAM         & 0.056          & 0.074          & 0.083          & 0.089          & 0.059          & 0.075          & 0.084          & 0.088          \\
		PAM         & \textbf{0.054} & \textbf{0.071} & \textbf{0.081} & \textbf{0.083} & \textbf{0.054} & \textbf{0.065} & \textbf{0.075} & \textbf{0.082} \\
		SPAM        & 0.055          & 0.072          & 0.083          & 0.088          & \textbf{0.054}          & 0.068          & 0.076          & 0.082          \\ \bottomrule
	\end{tabular}
	
	\label{ablation attention}
\end{table}

As shown in Table \ref{ablation attention}, PAM and SPAM are best and second best among all attention mechanisms tested regarding MSE. 
The utilization of the period router after obtaining the local information can dynamically extend receptive fields and identify periods of much longer scales. Therefore, PAM and SPAM can capture global dependencies among short periods and explore longer periods flexibly. 
LRAM, which discards prediction irrelevant segments by the low-rank assumption, achieves the third-best performance. 
LAM, which lacks cross-period modeling and may not efficiently aggregate global semantic information, exhibits a slightly higher prediction loss than LRAM. Vanilla full attention mechanism (FAM) suffers from temporal redundancy in time series and is easy to overfit to irrelevant interference, resulting in the worst performance. Based on the proposed period diffuser architecture, both LRAM and FAM outperform LogSparse Transformer and Informer, highlighting the effectiveness of our proposed architecture.

\subsubsection{Effect of iterative grouping mechanism}
In order to assess the relationship between the number of groups and the prediction loss, we conduct extensive experiments using PAM as the token mixer and the period diffuser as the predictor. These models are compared on the ETTh1, ETTh2, ETTm1, and ETTm2 datasets for comprehensive performance analysis. The input length of the encoder is fixed to 96, while the prediction length is set to 336. The experiments explore various numbers of groups ranging from 0 to 8.
For cases with the number of groups set to 0, we use traditional multivariate joint modeling techniques such as LogSparse Transformer, Informer, and Autoformer.
\begin{table}[h]
	\centering
	\caption{MSE results with different number of groups for MTS modeling on the ETT datasets.}
	\scriptsize
	\begin{tabular}{@{}c|cccc@{}}
		\toprule
		Num of Groups & ETTh1          & ETTh2          & ETTm1          & ETTm2          \\ \midrule
		0             & 0.505          & 0.438          & 0.430          & 0.329          \\
		1             & 0.468          & 0.431          & 0.400          & \textbf{0.300} \\
		2             & \textbf{0.442} & 0.416          & \textbf{0.393} & 0.305          \\
		4             & 0.460          & 0.414          & 0.401          & 0.315          \\
		8             & 0.475          & \textbf{0.411} & 0.402          & 0.329          \\ \bottomrule
	\end{tabular}
	
	\label{ablation grouping}
\end{table}

The models using IGM usually achieve much better performance than those employing traditional joint modeling (i.e., 0 group), as shown in Table \ref{ablation grouping}. Interestingly, each dataset tends to have an optimal number of groups that yields the best results. This highlights the intricate relationship among different time series within each dataset and provides insights into why traditional joint modeling approaches fall short.
In several cases, we observe that the prediction errors increase as the number of groups grows. This phenomenon can be attributed to the introduction of mismatched cross-variable dependencies resulting from a large number of groups.

By examining the positions of regrouping and ungrouping within the overall pipeline, we find that the feed-forward layer may be one of the key factors limiting the performance of vanilla Transformer-based models. Without the application of IGM, time series data related to different variables may become mixed, leading to serious confusion during the prediction of individual time series.

\subsubsection{Effect of period diffuser}
We compare the performance of different TSF predictors to unveil the potential of the period diffuser structure. To ensure a fair comparison, we maintain the same encoder structure based on PAM. All experiments are conducted on univariate time series data from the ETTh1 dataset, avoiding performance interference from various MTS modeling methods. In the experiments of the generative style decoder (GSD) and the fully connected network (FCN) predictors, we refer to the architectures of Informer and PatchTST for one-step prediction. In contrast, the period diffuser (PD) uses the features captured from long periods to short periods. We conduct experiments to forecast time series of different lengths, namely 96, 192, 336, and 720, for evaluating the performance of different predictors. Based on the results in Table \ref{ablation diffuser}, PD always achieves lower prediction losses compared to GSD and FCN across the target time series of different lengths, which demonstrates the effectiveness of the diffusion-style predictor in TSF and its ability to capture cross-period dependencies. The cross attention between each diffuser block and its counterpart encoder block in the period diffuser side effectively utilizes the extracted period information. The target time series are essentially generated in a progressive manner from long to short periods. Therefore, PD yields improved prediction accuracy compared with those obtained from traditional encoder-decoder architectures.
\begin{table}[h]
	\centering
	\caption{MSE results with three prediction structures on the ETTh1 dataset.}
	\scriptsize
	\begin{tabular}{@{}c|cccc|cccc@{}}
		\toprule
		Inp Len  & \multicolumn{4}{c|}{96}                                           & \multicolumn{4}{c}{336}                                  \\ \midrule
		Pred Len & 96             & 192            & 336            & 720            & 96             & 192            & 336            & 720   \\ \midrule
		PD       & \textbf{0.054} & \textbf{0.071} & \textbf{0.081} & \textbf{0.083} & \textbf{0.054} & \textbf{0.065} & \textbf{0.075} & \textbf{0.082} \\
		GSD      & 0.066          & 0.083          & 0.095          & 0.096          & 0.073          & 0.078          & 0.078          & 0.087 \\
		FCN      & 0.055          & 0.073          & 0.083          & 0.085          & 0.056          & 0.071          & 0.081          & 0.088 \\ \bottomrule
	\end{tabular}
	\label{ablation diffuser}
\end{table}

\section{Conclusion}
This paper proposes PeriodNet that incorporates period attention mechanism and sparse period attention mechanism to exploit the periodicity, locality and redundancy in time series forecasting. The proposed iterative grouping mechanism effectively extracts cross-variable dependencies while maintaining low computational overhead. The proposed period diffuser offers promising sensitivity to periods of various lengths, making it more suitable for time series forecasting tasks. Extensive numerical experiments demonstrate the excellent performance of our proposed models in both univariate and multivariate time series forecasting with respect to prediction loss. The proposed PeriodNet and SPeriodNet outperform six state-of-the-art models on multiple datasets in terms of MSE and MAE and they offer certain robustness in predicting time series of different lengths.

\section{Acknowledgements}
This work was partially supported by the Natural Science Foundation of Hebei Province (No. F2022105027) and the Fundamental Research Funds for the Central Universities, P. R. China.

 \bibliographystyle{elsarticle-num} 
\bibliography{refs}



\end{document}